%% file: main.tex
\documentclass[10pt,twocolumn,a4paper]{article}
\usepackage[margin=20mm]{geometry}
\usepackage{graphicx}
\usepackage{amsmath,amssymb}
\usepackage{booktabs,multirow,tabularx}
\usepackage[table]{xcolor}
\usepackage{float}
\usepackage{needspace}
\usepackage[font=small,labelfont=bf,hypcap=false]{caption}
\usepackage[numbers,sort&compress]{natbib}
\usepackage[hyphens]{url}
\usepackage[hidelinks]{hyperref}
\definecolor{rowgray}{gray}{0.95}
\newif\ifshowauthors
\showauthorstrue
\title{Multi-Rate Flow Matching for Point Cloud Upsampling\par\vspace{40pt}}
\author{Jingkai Shang\\
Institute for Artificial Intelligence\\
University of Stuttgart\\
Stuttgart, Germany\\
{\tt\small st189675@stud.uni-stuttgart.de}
\and
Edgar Torres\\
Institute for Artificial Intelligence\\
University of Stuttgart\\
Stuttgart, Germany\\
{\tt\small edgar.torres@ki.uni-stuttgart.de}
\and
Duy Nguyen\\
Institute for Artificial Intelligence\\
University of Stuttgart\\
Stuttgart, Germany \\
{\tt\small ho-minh-duy.nguyen@ki.uni-stuttgart.de}
\date{{\Large\textcolor{gray}{\textbf{Preprint}}}}
}

\begin{document}
\ifshowauthors\else\author{}\fi
\twocolumn[{%
  \begin{@twocolumnfalse}
    \maketitle
    \vspace{20pt}
    \begin{minipage}{\textwidth}
\centering
\includegraphics[width=\textwidth]{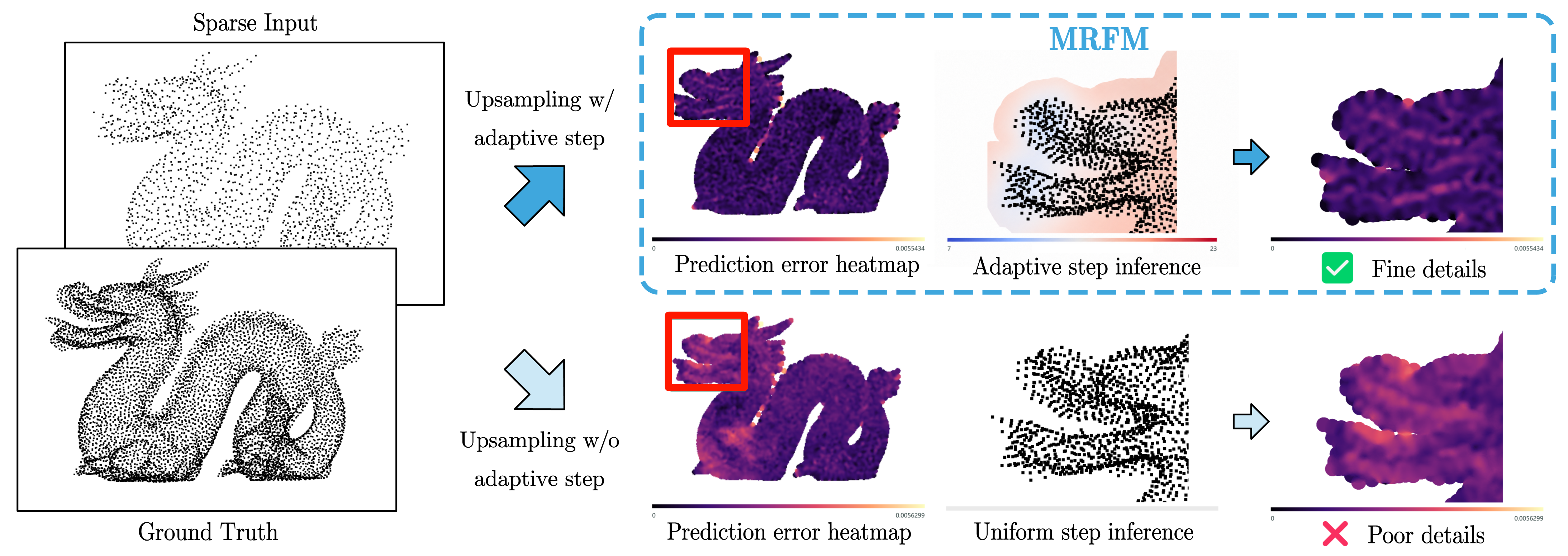}
\captionof{figure}{\textbf{Teaser.} We propose a Multi-Rate Flow Matching (MRFM) for point-cloud upsampling. Given the same sparse input, MRFM adaptively allocates more integration steps to difficult local regions, leading to lower global and local reconstruction errors than uniform-step inference while better preserving fine geometric details.}\label{fig:teaser}
    \end{minipage}\par
    \vspace{30pt}
\begin{abstract}
\noindent Flow-matching point-cloud upsamplers refine sparse inputs into dense point clouds by integrating a learned velocity field with a fixed temporal discretization. However, local reconstruction requirements vary substantially across a shape, while a shared solver budget applies the same temporal resolution everywhere. We propose Multi-Rate Flow Matching (MRFM), a patch-adaptive framework that selects an integration budget for each local patch at inference time while keeping the learned velocity field frozen. MRFM relaxes the discrete budget into a continuous variable, optimizes it online through differentiable soft rollouts using an input-derived geometric fidelity objective, and projects the result back to a discrete budget for the final solver rollout. Experiments on PU1K and PUGAN show strong reconstruction quality, while matched-budget permutation and multi-factor analyses demonstrate that the benefit depends on patch-specific allocation rather than on the step distribution alone. MRFM is accuracy-oriented and intentionally uses additional test-time optimization to obtain more suitable local temporal resolutions.
\end{abstract}
    \vspace{60pt}
  \end{@twocolumnfalse}
}]

\section{Introduction}

Point-cloud upsampling aims to recover dense, uniform, and geometrically faithful point sets from sparse observations. Sparse, noisy, and non-uniform point clouds are the default output of most 3D sensors, yet downstream tasks assume denser, more regular geometry: robotic mapping and SLAM need locally complete surfaces for reliable registration, LiDAR-based 3D detection for autonomous driving degrades on sparse or occluded returns, and AR/VR rendering needs smooth, hole-free surfaces \cite{li2019pugan}, \cite{qu2024pudm}, \cite{liu2026pufm}. The same sparse-to-dense problem also arises beyond 3D vision, for instance in sparse-view medical surface reconstruction or physical-field reconstruction from a limited sensor array. The problem is therefore not only to add points, but to infer plausible local geometry while preserving the structure the sparse observation already supports, particularly around boundaries, thin parts, and unevenly sampled regions where small local errors noticeably affect the reconstructed surface.

Early learned upsamplers such as PU-Net and its successors \cite{yu2018punet} expand sparse points in a learned feature space and substantially improve point distribution and geometric fidelity, but do so through one-shot or staged expansion rather than an explicit refinement trajectory. More recent methods instead treat upsampling as iterative or generative refinement -- optimizing point coordinates, denoising, or transporting a learned flow -- with PUFM \cite{liu2026pufm} and PUFM++ \cite{liu2025pufmpp} representing the current state of the art for flow-matching-based upsampling. These flow-based methods make the temporal discretization of the learned trajectory an explicit design choice, which motivates asking whether the same discretization is equally appropriate for every local region.

Local reconstruction difficulty, however, varies substantially within a single shape -- smooth, well-observed regions may need only a coarse trajectory, while boundaries, thin structures, and sparsely sampled neighborhoods require finer refinement -- yet a shared solver budget applies the same temporal resolution everywhere. The closest adaptive mechanisms address a different axis of this problem: PUFM++ redistributes a shared global time grid \cite{liu2025pufmpp}, Grad-PU applies gradients to point coordinates rather than the solver budget \cite{he2023gradpu}, PatchScaler groups image patches by diffusion difficulty \cite{liu2025patchscaler}, and TG-MSFM gates multi-scale velocity components over time \cite{wang2026tgmsfm}, but none of them selects a patch-specific flow-matching budget from the response of a frozen point-cloud upsampler. Even a spatially varying budget is not simply recoverable from the input geometry alone: curvature or density are informative descriptors, but they do not measure how the learned flow actually behaves on a particular patch, or whether that patch would benefit from additional refinement.

To address this gap, we propose Multi-Rate Flow Matching (MRFM), which assigns an individual integration budget to each local region. Inspired by gradient-guided generative optimization \cite{guo2024gradient}, MRFM relaxes the discrete budget into a continuous rate and optimizes it through differentiable soft rollouts using an input-derived geometric fidelity objective.

\begin{figure*}[!t]
\centering
\includegraphics[width=1.0\linewidth]{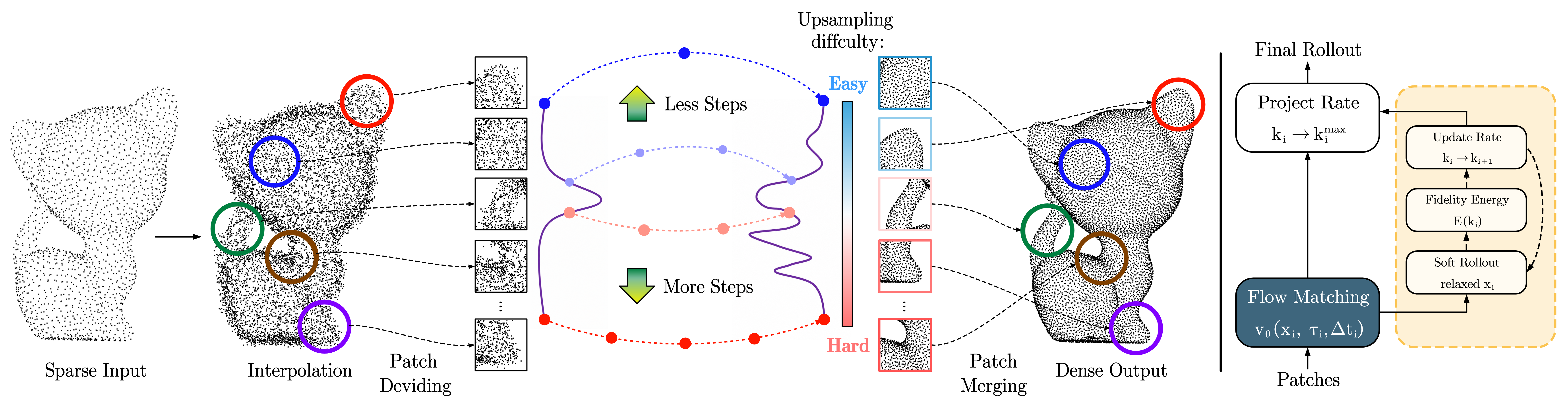}
\caption{\textbf{Overview of MRFM.} Left: the overall pipeline of the proposed framework. Right: the iterative gradient-guided rate optimization procedure during inference.}\label{fig:overall_structure}
\end{figure*}

The optimized rate is projected to an admissible discrete budget, and the final dense patch is recomputed with the standard flow solver. The velocity field remains fixed throughout. MRFM therefore uses additional test-time optimization to obtain patch-specific temporal resolutions and is designed for reconstruction accuracy rather than inference acceleration. The overall pipeline and the inference-time optimization loop are summarized in Figure~\ref{fig:overall_structure}.

Our main contributions are summarized as follows:

\begin{enumerate}
\item We introduce a patch-wise multi-rate formulation for flow-matching point-cloud upsampling, allowing local regions to use different integration budgets while sharing the same velocity model.

\item We develop a gradient-guided test-time rate optimization strategy that selects patch-specific budgets from the response of the frozen flow model, without a separately trained router or a hand-crafted allocation rule.

\item We evaluate MRFM on PU1K and PUGAN against state-of-the-art point-cloud upsampling methods, assessing reconstruction quality and computational cost. Controlled budget-permutation experiments and patch-level analyses further examine the benefit of spatial budget allocation and its relationship to reconstruction difficulty.

\end{enumerate}

\section{Related work}

\subsection{Point cloud upsampling}

Early learned upsamplers expand sparse points in a learned feature space and then generate new coordinates. PU-Net established this paradigm \cite{yu2018punet}, PU-GAN improved point uniformity with adversarial training \cite{li2019pugan}, 3PU progressively upsamples local patches \cite{yifan2019progressive}, PUGeo-Net emphasizes geometry-centric reconstruction \cite{qian2020pugeonet}, and PU-Transformer introduces self-attention \cite{qiu2022putransformer}. Although their architectures differ, these methods primarily learn how to represent local geometry and how to expand sparse features into denser coordinates. Their computation is therefore largely determined by the network design rather than by a patch-specific numerical refinement process. More recent methods increasingly use iterative generative refinement: Grad-PU follows gradients of a learned distance field \cite{he2023gradpu}, PUDM and PDANS employ conditional diffusion \cite{qu2024pudm}, \cite{zhang2025pdans}, and PUFM / PUFM++ use flow matching \cite{liu2026pufm}, \cite{liu2025pufmpp}. In these methods, reconstruction is expressed as a sequence of updates, making the number and placement of updates part of the inference procedure. MRFM builds directly on the flow-matching line but addresses a different dimension of adaptivity: spatially varying solver budgets across local patches.

\subsection{Flow matching and adaptive integration}

Flow matching and rectified flow learn continuous velocity fields along simple probability paths \cite{lipman2023flow}, \cite{liu2023rectified}. In point-cloud upsampling, PUFM uses EMD pre-alignment to obtain a short transport trajectory \cite{liu2026pufm}, while PUFM++ adapts the placement of evaluations on a shared inference time grid \cite{liu2025pufmpp}. These approaches improve the global transport path or its temporal sampling, but the same schedule remains shared by the full shape. Classical adaptive ODE solvers instead control numerical truncation error \cite{hairer1993solving}; this criterion is different from selecting computation according to reconstruction behavior, because local numerical residual and patch-level reconstruction quality need not coincide. Gradient-guided diffusion optimization is closer in spirit because it modifies an inference process using an external objective without retraining the generator \cite{guo2024gradient}. MRFM applies this optimization view to the temporal resolution of each patch while keeping $v_\theta$ frozen, so the learned transport itself is unchanged and only its local discretization is adapted.

\subsection{Multi-scale and patch-adaptive generative models}

Adaptive or multi-scale computation has also appeared outside point-cloud upsampling. TG-MSFM blends multi-scale velocity components with a time-dependent gate for time-series imputation, but the gating is global rather than patch-specific \cite{wang2026tgmsfm}. MLPCM organizes 3D generation across point- and super-point latent resolutions and uses consistency distillation for fast sampling \cite{du2024mlpcm}, while LaS-Comp couples latent and spatial consistency for zero-shot 3D completion \cite{yan2026lascomp}. PatchScaler is the closest computational analogue: it groups image patches by estimated reconstruction difficulty and assigns different diffusion sampling configurations \cite{liu2025patchscaler}. These approaches motivate adaptive computation at different scales, whereas MRFM specifically performs test-time, patch-wise solver-budget selection for flow-matching point-cloud upsampling.

\section{Method}

The overall framework is illustrated in Figure~\ref{fig:overall_structure}. Given a sparse point cloud, we first decompose it into overlapping local patches and expand each patch to the target cardinality using a deterministic interpolation operator. The expanded patches are then refined by a shared rate-conditioned flow model. Instead of using one uniform integration budget for all regions, MRFM optimizes a patch-specific continuous rate through gradient guidance at inference time. The optimized rates are projected to discrete budgets, after which the corresponding flow trajectories are recomputed and merged into the final dense point cloud.

\subsection{Multi-rate flow matching}

We formulate point-cloud upsampling as conditional flow matching over local geometric patches. In conventional flow matching, all regions evolve under shared continuous-time dynamics,

\begin{equation}
\frac{\mathrm{d}x}{\mathrm{d}t}=v_\theta(x,t),
\end{equation}

where $v_\theta$ denotes the learned velocity model. We decompose the sparse point cloud into patches $\mathcal{X}=\{x_i\}_{i=1}^M$ and interpolate each sparse patch $x_i$ to the target point count, obtaining the initial dense patch $x_i^0$.\footnote{This initialization is denoted by $H_i$ in Section 3.2.2.} For a budget $K_i\in\mathcal{K}=\{1,\ldots,K_{\mathrm{max}}\}$ and step $s\in\{0,\ldots,K_i-1\}$, define

\begin{equation}
\tau_i^s=\frac{s}{K_i},\quad\Delta t_i=\frac{1}{K_i}.
\end{equation}

The shared rate-conditioned velocity model is evaluated through the Euler update

\begin{equation}
x_i^{s+1}=x_i^s+\Delta t_i v_\theta(x_i^s,\tau_i^s,\Delta t_i).\label{eq:discrete-flow}
\end{equation}

Larger $K_i$ gives a finer trajectory, whereas smaller $K_i$ yields a coarser discretization. Importantly, all patches still traverse the same normalized interval and share the same frozen parameterized velocity model; only the numerical resolution changes. Hence MRFM adapts the solver budget rather than the learned transport itself. The remaining problem is to select $K_i$ from the sparse observation, as described in Section~\ref{sec:geometric-guidance}.

\subsection{Patch-level rate selection}

Local reconstruction requirements are heterogeneous: smooth, well-observed regions may need coarse trajectories, whereas boundaries, thin structures, sparse neighborhoods, or ambiguous geometry may benefit from finer resolution. We therefore define difficulty operationally through the response of the frozen flow model rather than a single predefined geometric statistic. Curvature or density can describe the input geometry, but neither directly measures whether the learned flow would benefit from additional refinement. MRFM instead adjusts each test-patch rate according to consistency between its generated dense patch and observed sparse geometry, coupling allocation to both input geometry and model-realized reconstruction behavior.

This distinction is important because the selected budget is not treated as a direct geometric label. Two patches with similar curvature or density can still receive different budgets when the frozen model produces different reconstruction responses. Conversely, a geometrically complex patch need not receive the maximum budget if its current trajectory is already consistent with the observed support. The rate therefore reflects the interaction between local geometry and the behavior of the trained flow model.

\subsubsection{Rate-conditioned velocity field}

The evolving coordinates $x_i$ are mapped by a geometric encoder to point-wise local features. A temporal/rate embedding from $(\tau_i^s,\Delta t_i)$ modulates the intermediate velocity representation before the shared predictor outputs the update in \eqref{eq:discrete-flow}. Thus, $\tau_i^s$ gives trajectory position and $\Delta t_i$ numerical resolution. The same frozen network operates across all budgets in $\mathcal{K}$; only the rate is adapted at test time. Conditioning on both quantities is important because the predictor must know not only where the current state lies on the trajectory but also the numerical resolution under which that state is advanced.

\subsubsection{Input-derived geometric guidance}\label{sec:geometric-guidance}

Let $G$ denote the observed sparse patch, $H$ its deterministically interpolated initialization/reference set, and $P$ the dense prediction produced by the current relaxed rollout. We combine sparse coverage, local surface consistency, and spacing regularization. In normalized patch coordinates, the guidance energy is

\begin{equation}
\begin{aligned}\mathcal{F}(P;G,H)={}&\mathcal{L}_{\mathrm{cov}}(P,G)\\&+\mathcal{L}_{\mathrm{surf}}(P,G)\\&+\mathcal{L}_{\mathrm{rep}}(P,H).\end{aligned}\label{eq:energy}
\end{equation}

All references come from the sparse input; dense ground truth, meshes, and ground-truth normals are not used. $H$ supplies initialization and a spacing reference rather than a dense supervision target. These constraints are used only for budget selection and remain distinct from the evaluation metrics used in Section 4.

\textbf{Sparse coverage.} Let $N_G$ denote the number of points in $G$. Each observed point should be represented in the generated patch, so we use the directed squared set discrepancy

\begin{equation}
\mathcal{L}_{\mathrm{cov}}(P,G)=\frac{1}{N_G}\sum_{g\in G}\min_{p\in P}\lVert g-p\rVert_2^2.\label{eq:coverage}
\end{equation}

This direction preserves measured support without attracting every generated point to sparse observations. A symmetric discrepancy to $G$ would also penalize valid intermediate samples simply because they lie between measurements, so the directed form is better suited to sparse-to-dense guidance.

\textbf{Local surface consistency.} Let $\mathcal{N}_g\subset\mathbb{R}^3$ denote the $K$ nearest sparse neighbors of $g$, where each $u\in\mathcal{N}_g$ is a neighboring sparse-point coordinate, $K$ is an implementation hyperparameter, and $K_g$ is the number of available neighbors (thus $K_g\le K$). We compute the neighborhood centroid $\mu_g$ and covariance matrix $\Sigma_g$ as

\begin{equation}
\mu_g=\frac{1}{K_g}\sum_{u\in\mathcal{N}_g}u,\label{eq:mean}
\end{equation}

\begin{equation}
\Sigma_g=\frac{1}{K_g}\sum_{u\in\mathcal{N}_g}(u-\mu_g)(u-\mu_g)^\top.\label{eq:covariance}
\end{equation}

The covariance matrix describes the spatial spread of the neighboring points. Assuming the surface is locally approximately planar, its normal is the direction of least variance. We therefore estimate $n_g\in\mathbb{R}^3$ as the unit eigenvector of $\Sigma_g$ associated with its smallest eigenvalue, equivalently solving

\begin{equation}
n_g\in\operatorname*{arg\,min}_{\lVert n\rVert_2=1}n^\top\Sigma_g n.\label{eq:normal}
\end{equation}

With $q(p)\in\operatorname*{arg\,min}_{g\in G}\lVert p-g\rVert_2^2$ and $N_P$ denoting the number of points in $P$, the point-to-plane energy is

\begin{equation}
\mathcal{L}_{\mathrm{surf}}(P,G)=\frac{1}{N_P}\sum_{p\in P}(n_{q(p)}^\top(p-q(p)))^2.\label{eq:surface}
\end{equation}

This residual constrains normal displacement while permitting tangential motion that can fill gaps between sparse observations. The squared projection is independent of normal orientation. Normals are estimated once from fixed observations; nearest-neighbor assignments are piecewise constant, so gradients exist almost everywhere. Local planes remain approximate near sharp edges or degenerate neighborhoods, which limits the reliability of this term in such regions.

\textbf{Spacing regularization.} Let the nearest-neighbor separation in the prediction and the reference spacing be

\begin{equation}
d_P(p)=\min_{p'\in P,\,p'\ne p}\lVert p-p'\rVert_2,\label{eq:spacing}
\end{equation}

\begin{equation}
\ell_H=\operatorname*{median}_{u\in H}\min_{u'\in H,\,u'\ne u}\lVert u-u'\rVert_2.\label{eq:scale}
\end{equation}

Writing $[z]_+=\max(z,0)$, the repulsion energy is

\begin{equation}
\mathcal{L}_{\mathrm{rep}}(P,H)=\frac{1}{N_P}\sum_{p\in P}[\ell_H-d_P(p)]_+^2.\label{eq:repulsion}
\end{equation}

This term suppresses clustering below the reference sampling scale and vanishes once that separation is reached, avoiding persistent outward expansion of the patch. Coverage anchors observed support, surface consistency controls normal displacement, and repulsion discourages local collapse; together they constrain complementary aspects of the same reconstruction without forcing agreement with interpolated coordinates.

\textbf{Differentiable rate relaxation.} We relax the integer budget to $\kappa_i\in[1,K_{\mathrm{max}}]$ and unroll $K_{\mathrm{max}}$ soft steps. For $s\in\{0,\ldots,K_{\mathrm{max}}-1\}$, define the smooth activity gate

\begin{equation}
a_i^s(\kappa_i)=\frac{1}{1+e^{-\alpha(\kappa_i-(s+1/2))}},\label{eq:gate}
\end{equation}

where $\alpha$ controls gate sharpness. For the relaxed rollout, set $\tau_i^s=s/\kappa_i$ and $\Delta t_i=1/\kappa_i$, with $\tilde{x}_i^0=H_i$. Then

\begin{equation}
\begin{aligned}\tilde{x}_i^{s+1}={}&\tilde{x}_i^s+a_i^s(\kappa_i)\Delta t_i\\&v_\theta(\tilde{x}_i^s,\tau_i^s,\Delta t_i),\end{aligned}\label{eq:soft-rollout}
\end{equation}

with endpoint $P_i(\kappa_i)=\tilde{x}_i^{K_{\mathrm{max}}}$. The activity gate smoothly controls how many updates contribute as $\kappa_i$ varies, enabling gradient-based rate selection while keeping the velocity network frozen.

\textbf{Optimization and discrete inference.} Using the guidance energy in \eqref{eq:energy}, for a batch $\mathcal{B}$ of $N_B$ patches, we minimize

\begin{equation}
\mathcal{J}(\boldsymbol{\kappa})=\frac{1}{N_B}\sum_{i\in\mathcal{B}}\mathcal{F}(P_i(\kappa_i);G_i,H_i).\label{eq:batch}
\end{equation}

Let $\mathcal{F}_i=\mathcal{F}(P_i(\kappa_i);G_i,H_i)$ and define the endpoint sensitivity $S_i^{K_{\mathrm{max}}}=\partial P_i/\partial\kappa_i$. The rate gradient is

\begin{equation}
\frac{\partial\mathcal{J}}{\partial\kappa_i}=\frac{1}{N_B}\operatorname{tr}\bigl((\nabla_{P_i}\mathcal{F}_i)^\top S_i^{K_{\mathrm{max}}}\bigr),\label{eq:gradient}
\end{equation}

where the trace is the Frobenius inner product over point-coordinate derivatives. Network parameters, geometric features, and all reference sets remain fixed throughout this optimization; only $\boldsymbol{\kappa}$ is updated and projected onto $[1,K_{\mathrm{max}}]$. The optimizer, initialization, learning rate, and number of guidance iterations are implementation choices reported separately rather than part of the formulation.

After optimization, we project each continuous rate to the nearest admissible integer budget,

\begin{equation}
\hat{K}_i=\operatorname*{arg\,min}_{K\in\mathcal{K}}|K-\kappa_i|.\label{eq:projection}
\end{equation}

We restart the standard Euler rollout from $H_i$ with $\Delta t_i=1/\hat{K}_i$, transform predictions back to original coordinates, and merge overlapping outputs. No gradient-guided coordinate update is applied to the final prediction: only the budget is optimized. The relaxation and projection are approximate, and their forward/backward passes add cost beyond the final rollout.

\begin{table*}[p]
\centering
\caption{\textbf{Quantitative comparison with state-of-the-art methods.}
We report mean $\pm$ std. for CD ($10^{-4}$), HD ($10^{-2}$),
P2F ($10^{-2}$), and JSD ($10^{-1}$) on PUGAN and PU1K.}
\label{tab:main-results}

\setlength{\tabcolsep}{3pt}
\renewcommand{\arraystretch}{1.08}
\fontsize{9}{10.8}\selectfont

\newcommand{\mstd}[2]{%
  \shortstack[c]{%
    #1\\[-1.1pt]
    {\fontsize{6.5}{6.8}\selectfont\ttfamily
    $\pm$\texttt{#2}}%
  }%
}

\begin{tabularx}{\textwidth}{@{}cc*{8}{>{\centering\arraybackslash}X}@{}}
\toprule

\multirow{2}{*}{\shortstack{Upsampling\\rate}}
& \multirow{2}{*}{Method}
& \multicolumn{4}{c}{PU1K}
& \multicolumn{4}{c}{PUGAN} \\

\cmidrule(lr){3-6}
\cmidrule(lr){7-10}

& & CD & HD & P2F & JSD
& CD & HD & P2F & JSD \\

\midrule

% ========================= x4 =========================

\multirow{8}{*}{$\times4$}

& PU-GCN
& \mstd{5.1497}{0.0412}
& \mstd{6.9526}{0.0695}
& \mstd{1.7702}{0.0124}
& \mstd{2.9957}{0.0300}
& \mstd{11.404}{0.0912}
& \mstd{8.4747}{0.0848}
& \mstd{2.3706}{0.0166}
& \mstd{2.8373}{0.0284} \\

& Grad-PU
& \mstd{358.02}{5.3703}
& \mstd{6.5865}{0.0659}
& \mstd{1.5778}{0.0158}
& \mstd{2.8070}{0.0337}
& \mstd{348.18}{5.2227}
& \mstd{6.4313}{0.0643}
& \mstd{1.5865}{0.0159}
& \mstd{2.6951}{0.0323} \\

& PUDM
& \mstd{4.9152}{0.0590}
& \mstd{6.2694}{0.0878}
& \mstd{1.7081}{0.0222}
& \mstd{2.9511}{0.0354}
& \mstd{10.781}{0.1294}
& \mstd{9.7364}{0.1363}
& \mstd{2.4151}{0.0314}
& \mstd{2.6995}{0.0324} \\

& PDANS
& \mstd{4.5811}{0.0596}
& \mstd{6.3328}{0.0950}
& \mstd{1.6712}{0.0234}
& \mstd{2.7838}{0.0362}
& \mstd{10.389}{0.1351}
& \mstd{9.7406}{0.1461}
& \mstd{2.3345}{0.0327}
& \mstd{2.5931}{0.0337} \\

& RepKPU
& \mstd{4.5694}{0.0366}
& \mstd{5.4980}{0.0550}
& \mstd{1.5626}{0.0125}
& \mstd{2.7861}{0.0279}
& \mstd{10.969}{0.0878}
& \mstd{8.6813}{0.0868}
& \mstd{2.2229}{0.0178}
& \mstd{2.7421}{0.0274} \\

& PUFM
& \mstd{4.4648}{0.0268}
& \mstd{6.2605}{0.0501}
& \mstd{1.6675}{0.0117}
& \mstd{2.7408}{0.0219}
& \mstd{10.098}{0.0606}
& \mstd{9.6718}{0.0774}
& \mstd{2.2720}{0.0159}
& \mstd{2.5072}{0.0201} \\

& PUFM++
& \mstd{4.1887}{0.0209}
& \mstd{6.1932}{0.0434}
& \mstd{1.6167}{0.0097}
& \mstd{2.6230}{0.0184}
& \mstd{10.066}{0.0503}
& \mstd{9.7018}{0.0679}
& \mstd{2.2567}{0.0135}
& \mstd{2.4958}{0.0175} \\

\rowcolor{rowgray}
& MRFM (Ours)
& \mstd{4.0434}{0.0213}
& \mstd{6.2394}{0.0499}
& \mstd{1.5498}{0.0109}
& \mstd{2.4754}{0.0198}
& \mstd{10.044}{0.0703}
& \mstd{9.5972}{0.0768}
& \mstd{2.2554}{0.0158}
& \mstd{2.4838}{0.0199} \\

\midrule

% ========================= x8 =========================

\multirow{8}{*}{$\times8$}

& PU-GCN
& \mstd{4.4839}{0.0413}
& \mstd{10.606}{0.1220}
& \mstd{2.0550}{0.0165}
& \mstd{2.4876}{0.0286}
& \mstd{10.955}{0.1008}
& \mstd{12.539}{0.1442}
& \mstd{3.0876}{0.0249}
& \mstd{2.6718}{0.0307} \\

& Grad-PU
& \mstd{309.48}{5.3385}
& \mstd{7.2948}{0.0839}
& \mstd{1.8929}{0.0218}
& \mstd{1.9795}{0.0273}
& \mstd{306.30}{5.2837}
& \mstd{7.2232}{0.0831}
& \mstd{1.8760}{0.0216}
& \mstd{1.9395}{0.0268} \\

& PUDM
& \mstd{4.3965}{0.0607}
& \mstd{7.9176}{0.1275}
& \mstd{1.9757}{0.0295}
& \mstd{2.3339}{0.0322}
& \mstd{10.305}{0.1422}
& \mstd{12.163}{0.1958}
& \mstd{3.0493}{0.0456}
& \mstd{2.1857}{0.0302} \\

& PDANS
& \mstd{4.2394}{0.0634}
& \mstd{7.2659}{0.1253}
& \mstd{1.9639}{0.0316}
& \mstd{2.2045}{0.0330}
& \mstd{10.163}{0.1519}
& \mstd{12.359}{0.2132}
& \mstd{3.0210}{0.0486}
& \mstd{2.0762}{0.0310} \\

& RepKPU
& \mstd{4.1005}{0.0377}
& \mstd{6.6605}{0.0766}
& \mstd{1.9874}{0.0183}
& \mstd{2.6182}{0.0301}
& \mstd{10.583}{0.0974}
& \mstd{11.029}{0.1268}
& \mstd{1.7530}{0.0161}
& \mstd{2.8294}{0.0325} \\

& PUFM
& \mstd{3.9206}{0.0271}
& \mstd{7.2191}{0.0664}
& \mstd{1.9202}{0.0155}
& \mstd{2.0957}{0.0193}
& \mstd{9.6827}{0.0668}
& \mstd{12.388}{0.1140}
& \mstd{2.9331}{0.0236}
& \mstd{1.9605}{0.0180} \\

& PUFM++
& \mstd{3.6894}{0.0212}
& \mstd{7.3129}{0.0589}
& \mstd{1.8822}{0.0130}
& \mstd{1.9505}{0.0157}
& \mstd{9.7085}{0.0558}
& \mstd{12.375}{0.0996}
& \mstd{2.9350}{0.0203}
& \mstd{1.9610}{0.0158} \\

\rowcolor{rowgray}
& MRFM (Ours)
& \mstd{3.6505}{0.0214}
& \mstd{7.2031}{0.0663}
& \mstd{1.8478}{0.0149}
& \mstd{1.8603}{0.0171}
& \mstd{9.6855}{0.0780}
& \mstd{10.239}{0.0942}
& \mstd{2.9295}{0.0236}
& \mstd{1.9571}{0.0180} \\

\midrule

% ========================= x16 =========================

\multirow{8}{*}{$\times16$}

& PU-GCN
& \mstd{4.2235}{0.0439}
& \mstd{14.756}{0.1918}
& \mstd{2.7946}{0.0254}
& \mstd{2.5059}{0.0326}
& \mstd{6.9081}{0.0718}
& \mstd{8.5655}{0.1114}
& \mstd{2.6782}{0.0244}
& \mstd{2.9566}{0.0384} \\

& Grad-PU
& \mstd{220.11}{4.2921}
& \mstd{7.1082}{0.0924}
& \mstd{2.1218}{0.0276}
& \mstd{1.7805}{0.0278}
& \mstd{185.63}{3.6202}
& \mstd{7.0942}{0.0922}
& \mstd{1.9828}{0.0258}
& \mstd{1.3839}{0.0216} \\

& PUDM
& \mstd{3.3917}{0.0529}
& \mstd{7.8978}{0.1437}
& \mstd{2.1330}{0.0360}
& \mstd{2.1151}{0.0330}
& \mstd{6.8080}{0.1062}
& \mstd{12.227}{0.2225}
& \mstd{3.2706}{0.0553}
& \mstd{1.8765}{0.0293} \\

& PDANS
& \mstd{3.3401}{0.0564}
& \mstd{7.2083}{0.1406}
& \mstd{2.1118}{0.0384}
& \mstd{1.9859}{0.0336}
& \mstd{6.5532}{0.1108}
& \mstd{12.412}{0.2420}
& \mstd{3.1942}{0.0581}
& \mstd{1.6133}{0.0273} \\

& RepKPU
& \mstd{2.9452}{0.0306}
& \mstd{6.5117}{0.0847}
& \mstd{2.2273}{0.0232}
& \mstd{2.4315}{0.0316}
& \mstd{6.3546}{0.0661}
& \mstd{11.549}{0.1501}
& \mstd{1.8365}{0.0191}
& \mstd{1.7722}{0.0230} \\

& PUFM
& \mstd{2.4902}{0.0194}
& \mstd{7.3696}{0.0766}
& \mstd{2.0364}{0.0185}
& \mstd{1.8210}{0.0189}
& \mstd{5.6221}{0.0439}
& \mstd{12.155}{0.1264}
& \mstd{3.1224}{0.0284}
& \mstd{1.5341}{0.0160} \\

& PUFM++
& \mstd{2.2119}{0.0144}
& \mstd{7.1033}{0.0646}
& \mstd{1.9847}{0.0155}
& \mstd{1.4091}{0.0128}
& \mstd{5.2468}{0.0341}
& \mstd{11.871}{0.1080}
& \mstd{3.0609}{0.0239}
& \mstd{1.2574}{0.0114} \\

\rowcolor{rowgray}
& MRFM (Ours)
& \mstd{1.6346}{0.0139}
& \mstd{6.1206}{0.0637}
& \mstd{1.5663}{0.0143}
& \mstd{1.0670}{0.0111}
& \mstd{4.1354}{0.0376}
& \mstd{9.4938}{0.0987}
& \mstd{2.3634}{0.0215}
& \mstd{1.0646}{0.0111} \\

\bottomrule
\end{tabularx}
\end{table*}

\section{Experiments}

\subsection{Experiment Setup}

\textbf{Dataset.} We evaluate on the PUGAN \cite{li2019pugan} and PU1K \cite{qian2021pugcn} benchmarks, which are widely used for evaluating learned point-cloud upsampling methods. We use the official training splits and sparse test inputs containing 256 points, and report 4$\times$ and 8$\times$ upsampling, with 16$\times$ included as an additional evaluation setting in Table 1. Evaluating on both datasets allows us to examine the same rate-selection mechanism under different shape collections and output cardinalities without changing the underlying experimental protocol.

\textbf{Evaluation Metrics.} Reconstruction quality is measured by Chamfer Distance (CD), Hausdorff Distance (HD), Point-to-Surface Distance (P2F), and Jensen-Shannon Divergence (JSD), following standard point-cloud upsampling protocols \cite{li2019pugan}, \cite{qian2021pugcn}, \cite{liu2026pufm}. CD measures average set-level proximity, HD emphasizes the largest geometric discrepancy, P2F evaluates consistency with the underlying surface, and JSD reflects distributional agreement. We report the same metric definitions and scaling for all compared methods.

\textbf{Implementation Details.} The model is trained for 40 epochs with AdamW, batch size 64, and an initial learning rate of $1\times10^{-4}$. All experiments are conducted on a single NVIDIA GeForce RTX 2080 Ti GPU with 11 GB memory.

\begin{figure*}[p]
\centering
\includegraphics[width=0.97\linewidth]{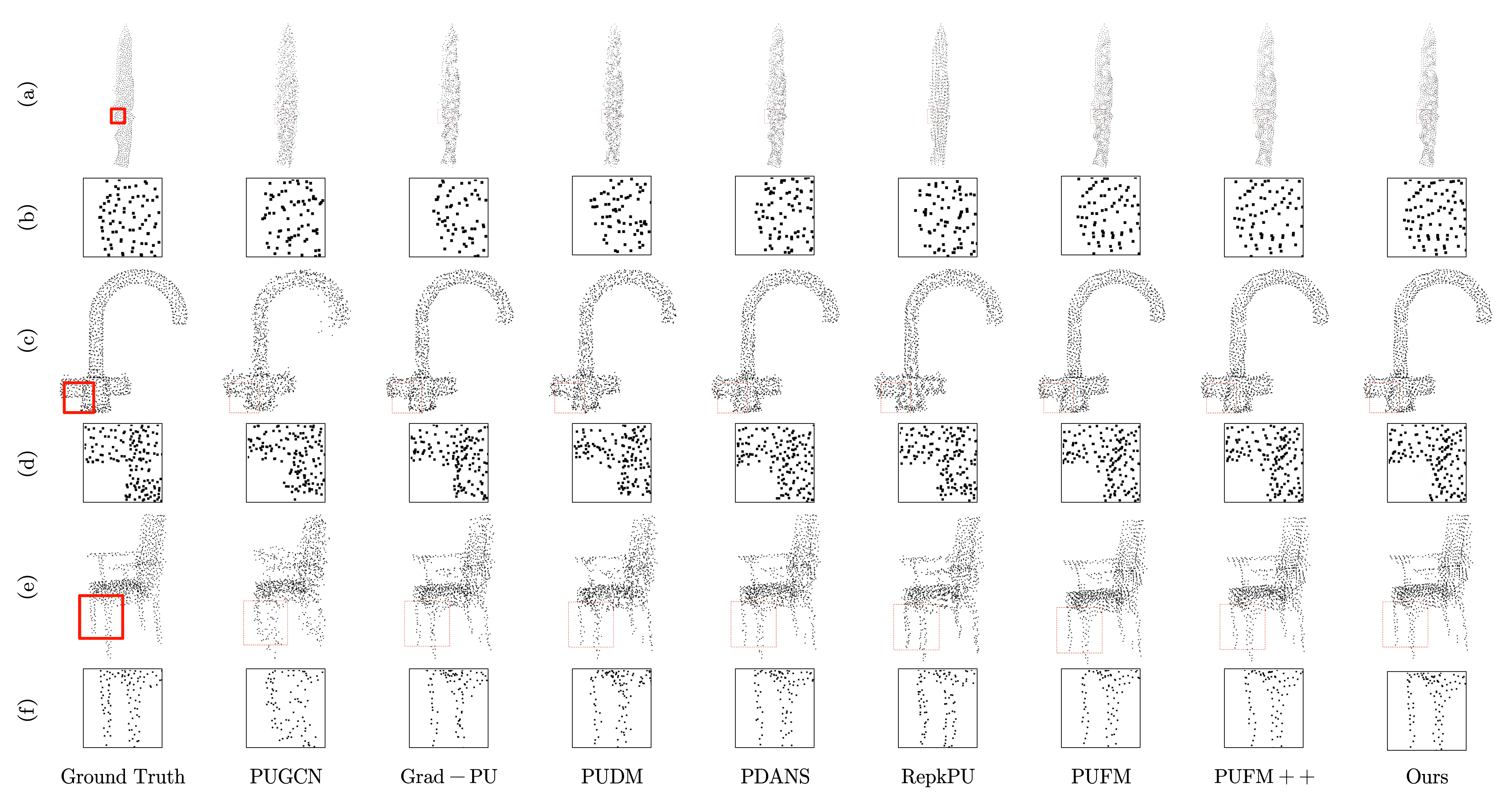}
\caption{\textbf{Visual comparison on 4$\times$ upsampling.} (a)-(b) smooth surfaces, (c)-(d) high-curvature structures, and (e)-(f) fine-grained structures; (b), (d), and (f) show the corresponding enlarged regions.}\label{fig:comparison}
\end{figure*}

\subsection{Main Results}

\begin{figure*}[!t]
\centering
\includegraphics[width=1.0\linewidth]{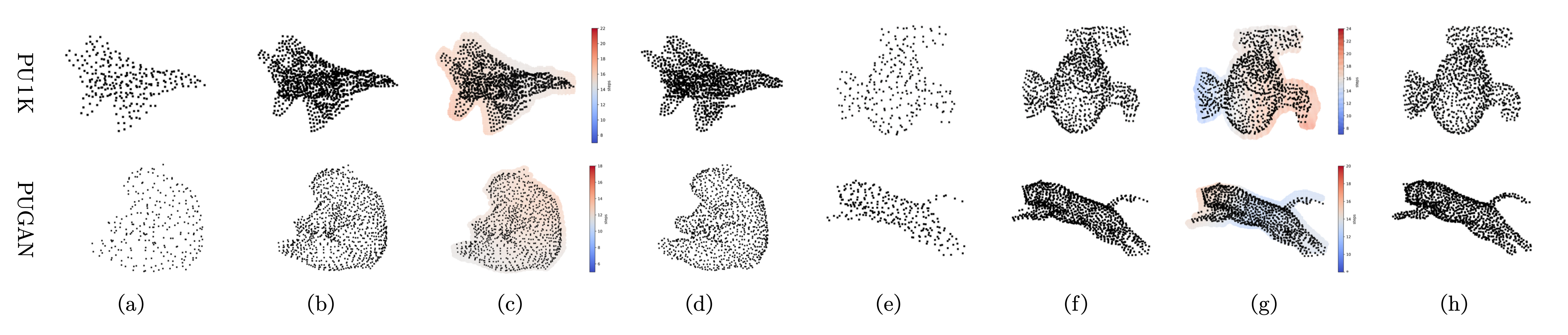}
\caption{\textbf{Visual demonstration of patch-wise step allocation.} (a) and (e) sparse input, (b) and (f) prediction, (c) and (g) step allocation map, and (d) and (h) ground truth. Color scales are shown separately for the two shapes.}\label{fig:step_allocation}
\end{figure*}

\textbf{Comparison with State-of-the-Art.} We compare MRFM with representative point-cloud upsampling methods, including PU-GCN \cite{qian2021pugcn}, Grad-PU \cite{he2023gradpu}, PUDM \cite{qu2024pudm}, PDANS \cite{zhang2025pdans}, RepKPU \cite{rong2024repkpu}, PUFM \cite{liu2026pufm}, and PUFM++ \cite{liu2025pufmpp}, using publicly released models under a unified evaluation framework. As shown in Table~\ref{tab:main-results}, MRFM achieves strong performance on both PU1K and PUGAN under 4$\times$ and 8$\times$ upsampling; the table also includes the 16$\times$ setting. On PU1K it obtains the best CD, P2F, and JSD at both ratios, including a CD of 4.0434 at 4$\times$ and 3.6505 at 8$\times$. Relative to the direct flow-matching baselines, the consistent gains across these metrics indicate that adapting the local integration budget provides complementary benefit beyond using a shared flow trajectory alone. On PUGAN, MRFM achieves the best CD and JSD at 4$\times$ and the best HD at 8$\times$, while remaining competitive on the other metrics. The gains are therefore not confined to a single metric or a single upsampling ratio, although their magnitude varies across datasets. The qualitative examples in Figure~\ref{fig:comparison} further show improved surface regularity and preservation of high-curvature and fine-grained structures. These visual differences are consistent with the quantitative results: the benefit is most apparent in local regions where inaccurate point placement or insufficient refinement affects the reconstructed geometry. The table is a reconstruction-quality comparison; because MRFM introduces an additional test-time guidance stage, these results should not be interpreted as evidence of inference acceleration.

\begin{figure}[!t]
\centering
\includegraphics[width=0.94\linewidth]{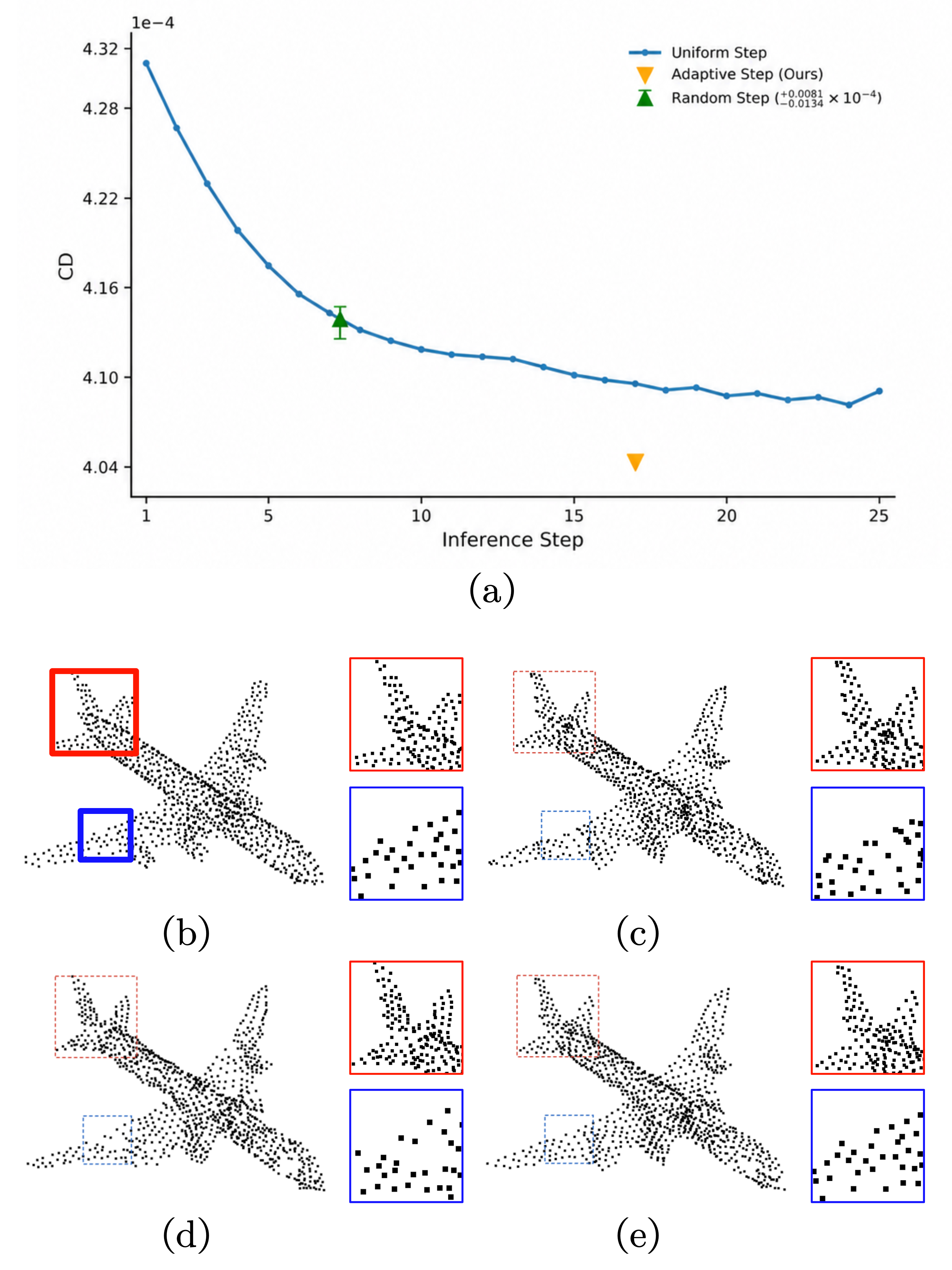}
\caption{\textbf{Effect of patch-wise adaptive inference on PU1K.} (a) CD versus uniform inference steps, with random-permutation and MRFM allocations shown for comparison. (b) Ground truth. (c-d) Results using 5 and 20 uniform inference steps. (e) MRFM adaptive-step result.}\label{fig:ablation}
\end{figure}

\textbf{Patch-wise Step Allocation.} We next visualize how MRFM distributes integration steps across local regions. As shown in Figure~\ref{fig:step_allocation}, the allocation is clearly non-uniform and varies within each shape, confirming that the optimization does not collapse to a single global rate. Higher-curvature and sparser patches often receive larger budgets, but the pattern is not explained by a single geometric cue; visually similar regions can still receive different budgets when their reconstruction responses differ. This is consistent with the design of MRFM, which never directly thresholds curvature, density, or surface statistics when assigning steps. Instead, these geometric properties can influence the generated trajectory, while the final rate is selected through the input-consistency response of the frozen flow model. Section 4.3 quantifies this relationship with both geometric factors and fixed-budget reconstruction error, complementing the qualitative allocation maps shown here.

\subsection{Ablation Studies}

\textbf{Allocation Effectiveness.} We evaluate the same frozen flow model with uniform inference budgets from 1 to 25 steps. As shown in Figure~\ref{fig:ablation}(a), reconstruction quality improves rapidly at small budgets and then saturates, indicating that simply increasing a shared step count yields diminishing returns. To isolate the effect of allocation from the amount of final solver computation, we randomly permute the patch-wise budgets predicted by MRFM. This preserves exactly the same budget multiset, average budget of 16.48 steps per patch, and total final-rollout Patch-NFE; only the patch-to-budget assignment changes. We repeat the permutation over 10 random reshuffles to avoid drawing conclusions from a single assignment. The randomly permuted allocation performs worse than the original MRFM allocation, showing that the gain depends on where the available solver effort is assigned rather than merely on using heterogeneous budgets or a larger average budget. The control is therefore deliberately strict: MRFM and the random baseline use the same collection of discrete step counts, and only their spatial correspondence differs. The qualitative examples in Figure~\ref{fig:ablation}(b-e) provide the corresponding local comparison and illustrate the same allocation behavior visualized in Figure~\ref{fig:step_allocation}.

\textbf{Multi-Factor Allocation Analysis.} To examine what is associated with MRFM's patch-wise allocation, we select 500 patches with diverse local structures from the PU1K and PUGAN test sets and compute Spearman $\rho$ between the predicted steps and several patch-level factors. We consider curvature, sparsity, surface non-uniformity, and model-realized reconstruction difficulty measured by CD@8 and P2F@8 under a common 8-step budget. The fixed-budget errors provide a common reference because every patch is evaluated with the same solver budget before correlation is measured. Random permutations preserve the same step set, mean budget, and total Patch-NFE, providing a control for correlations that could arise from the marginal step distribution alone. As shown in Table~\ref{tab:factor_correlation}, CD@8 has the strongest association with MRFM steps ($\rho=0.4812$), followed by curvature ($\rho=0.3161$) and sparsity ($\rho=0.2575$). P2F@8 is weaker ($\rho=0.1412$), while surface non-uniformity has only a limited association ($\rho=0.0602$). In contrast, the random-permutation correlations remain near zero across all factors. The different correlation strengths indicate that the allocation cannot be reduced to a single geometric rule: static properties such as curvature and sparsity are informative, but the strongest association is observed for the reconstruction error realized by the frozen flow under a common budget. Together, these results support a multi-factor interpretation of the predicted rates, in which geometry and model behavior are both reflected in the patch-specific allocation.
\begin{table}[!t]
\centering
\caption{\textbf{Multi-factor allocation analysis.} Spearman $\rho$ between MRFM-predicted steps and patch-level factors. Random-permutation results are reported as mean $\pm$ std.}\label{tab:factor_correlation}
\setlength{\tabcolsep}{3pt}
\renewcommand{\arraystretch}{1.12}
\footnotesize
\begin{tabular}{ccc}
\toprule
Factor & MRFM step $\rho$ & Random perm. $\rho$ \\
\midrule
Curvature & 0.3161 & 0.0191$\pm$0.0451 \\
Sparsity & 0.2575 & $\approx$0 \\
Surface non-uniformity & 0.0602 & 0.0154$\pm$0.0395 \\
CD@8 & 0.4812 & 0.0197$\pm$0.0432 \\
P2F@8 & 0.1412 & 0.0134$\pm$0.0482 \\
\bottomrule
\end{tabular}
\end{table}

Taken together, the matched-budget permutation and multi-factor analyses provide two complementary views of the same behavior. The permutation experiment shows that preserving the step distribution while destroying the patch-to-budget correspondence reduces reconstruction quality, whereas the correlation analysis shows that the original correspondence is structured rather than random. MRFM therefore benefits from the assignment itself, and that assignment reflects several complementary patch properties rather than a single geometric heuristic. Together with the permutation control, they support interpreting the selected rate through the response of the trained model.

\section{Conclusion}

We presented Multi-Rate Flow Matching (MRFM), a patch-adaptive framework for point-cloud upsampling that selects local integration budgets through gradient-guided optimization of a frozen flow model. Experiments on PU1K and PUGAN show strong reconstruction quality, while controlled ablations demonstrate that gains arise from patch-specific allocation rather than the step distribution alone. The selected rates reflect both geometric and model-realized difficulty. MRFM uses additional test-time optimization and is therefore accuracy-oriented rather than an acceleration method. Its sparse-coverage, local-surface, and spacing objectives guide budget selection without modifying the frozen velocity model, while final predictions are recomputed using discrete budgets.

The central distinction is between adapting a shared temporal schedule and adapting the amount of refinement assigned to individual spatial regions. MRFM uses the response of the frozen model to select the latter, rather than treating curvature or density as a direct prescription for computation. The continuous relaxation makes this selection differentiable, while projection and a restarted discrete rollout preserve a conventional final integration procedure. The resulting mechanism changes the local numerical resolution without requiring a separately trained allocation network or a gradient update to the final point coordinates.

The matched-budget permutation experiment is particularly informative in this setting. Preserving the same collection of step counts while changing their spatial assignment reduces reconstruction quality, showing that heterogeneous budgets alone do not explain the improvement. The patch-level analysis complements this result: geometric factors are associated with the selected rates, but their correlations differ, and fixed-budget reconstruction error provides the strongest association among the evaluated factors. These observations support interpreting allocation through the interaction between observed geometry and the behavior of the learned flow.

Soft rollouts and online optimization add cost, and integer projection remains approximate. Local planes can be unreliable near sharp edges or degenerate neighborhoods. Future work should reduce guidance overhead and improve geometric reliability.

\clearpage
{\small
\bibliographystyle{ieeenat_fullname}
\bibliography{main}
}
\clearpage
\onecolumn
\appendix

\input{appendix}
\end{document}

%% file: appendix.tex
\section{Implementation details}

PointNet++ is adopted as the backbone for feature extraction, while Mamba blocks are used as the velocity predictor in the flow-matching model. The remaining implementation details are summarized in Table~\ref{tab:implementation_details}.

\begin{table}[H]
\centering
\caption{\textbf{Other implementation details.}}\label{tab:implementation_details}
\setlength{\tabcolsep}{7pt}
\renewcommand{\arraystretch}{1.18}
\small
\begin{tabular}{cc}
\toprule
Hyperparameter & Value \\
\midrule
Patch size & 64 \\
Surface KNN neighbors & 12 \\
Mamba layers & 6 \\
Time embedding dimension & 32 \\
Mamba state dimension & 32 \\
Mamba convolution width & 4 \\
Mamba expansion factor & 4 \\
Integration budget range & $1$--$25$ \\
Initial step budget & 13 \\
Maximum guidance iterations & 12 \\
Guidance learning rate & 1.0 \\
\bottomrule
\end{tabular}
\end{table}

\section{Additional results}

\Needspace{8\baselineskip}
\subsection{Computational Efficiency Analysis}

We further compare the computational cost of MRFM with the methods evaluated in the main state-of-the-art comparison. All methods are evaluated under the same hardware and input setting. We report inference latency and peak GPU memory for all methods. For iterative flow-based methods, we additionally report the average number of function evaluations per input instance; for MRFM, this corresponds to the average number of refinement steps per patch. A dash indicates that the corresponding metric is not applicable to the inference mechanism of that method.

\begin{table}[H]
\centering
\caption{\textbf{Computational cost comparison with state-of-the-art methods.}}\label{tab:sota_compute}
\setlength{\tabcolsep}{7pt}
\renewcommand{\arraystretch}{1.18}
\small
\begin{tabular}{cccc}
\toprule
Method & NFE & Latency (ms) & Peak Memory (GB) \\
\midrule
PU-GCN & 3 & 113.1692 & 0.1401 \\
Grad-PU & 33 & 276.3827 & 0.1444 \\
PUDM & 10 & 2384.2558 & 0.2223 \\
PDANS & 9 & 2275.4462 & 0.7275 \\
RepKPU & 3 & 144.9831 & 0.1444 \\
PUFM & 5 & 871.4086 & 0.2050 \\
PUFM++ & 5 & 205.2651 & 0.1681 \\
\cellcolor{rowgray}MRFM (Ours) & \cellcolor{rowgray}16.48 & \cellcolor{rowgray}936.1042 & \cellcolor{rowgray}0.1413 \\
\bottomrule
\end{tabular}
\end{table}

As reported in Table~\ref{tab:sota_compute}, MRFM deliberately introduces additional test-time computation to improve reconstruction quality. Its average NFE reaches 16.48, compared with 5 for PUFM and PUFM++, reflecting the extra refinement enabled by patch-wise adaptive integration and gradient-guided rate optimization. This additional computation results in increased inference latency, while the peak GPU memory remains comparable to or lower than that of most competing methods. These results are consistent with the accuracy-oriented design of MRFM: the method intentionally trades additional inference-time computation for improved reconstruction quality rather than targeting inference acceleration.

\Needspace{8\baselineskip}
\subsection{Alternative Solver Comparisons}

We compare MRFM with two alternative configurations: a pure multi-layer perceptron (MLP) model and the classical adaptive Runge--Kutta solver (RK45). We replace gradient-guided step allocation with the MLP and the RK45, respectively. We test the three configurations using Chamfer Distance (CD), Hausdorff Distance (HD), Point-to-Surface Distance (P2F), and Jensen-Shannon Divergence (JSD). The quantitative results are summarized in Table~\ref{tab:alternative-solvers}.
\begin{table}[H]
\centering
\caption{\textbf{Comparison with alternative solver.}
    We report CD ($10^{-4}$), HD ($10^{-2}$),
    P2F ($10^{-2}$), and JSD ($10^{-1}$).}\label{tab:alternative-solvers}
\setlength{\tabcolsep}{7pt}
\renewcommand{\arraystretch}{1.18}
\small
\begin{tabular}{ccccc}
\toprule
Method & CD & HD & P2F & JSD \\
\midrule
MLP & 4.0863 & 6.2755 & 1.5506 & 2.5730 \\
RK45 & 4.0712 & 6.2502 & 1.5501 & 2.5128 \\
\cellcolor{rowgray}MRFM (Ours) & \cellcolor{rowgray}4.0434 & \cellcolor{rowgray}6.2394 & \cellcolor{rowgray}1.5498 & \cellcolor{rowgray}2.4754 \\
\bottomrule
\end{tabular}
\end{table}

As shown in Table~\ref{tab:alternative-solvers}, MRFM achieves the best results across all four metrics, outperforming both MLP and RK45. This shows that reconstruction-guided patch-wise rate allocation is more effective than learned or classical adaptive alternatives.

\Needspace{8\baselineskip}
\subsection{Downstream Application}

\begin{figure}[H]
\centering
\includegraphics[width=0.85\linewidth]{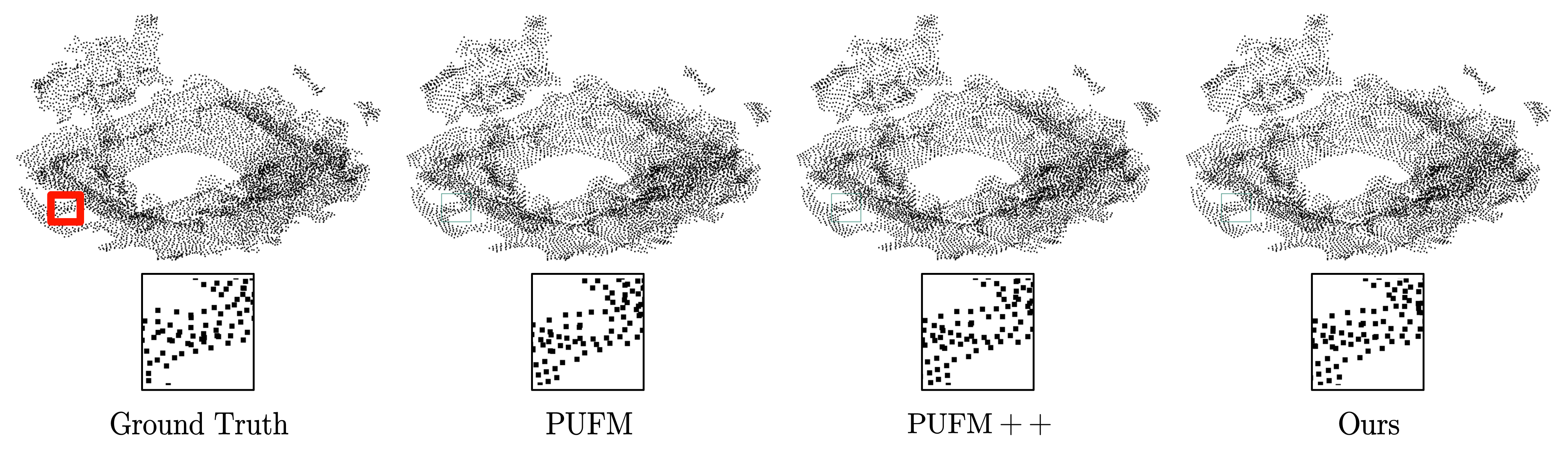}
\caption{\textbf{Qualitative comparison on ScanNet.}}\label{fig:scannet_realworld}
\end{figure}

We also evaluate the upsampling performance of our
method on real-world data. We test MRFM together with two strong flow-matching baselines, PUFM and PUFM++, on ScanNet. We report Chamfer Distance (CD, $10^{-3}$) as the quantitative metric. PUFM obtains a CD of 1.4490, PUFM++ improves slightly to 1.2305, while MRFM achieves the best result of 1.2112. The qualitative comparison is shown in Figure~\ref{fig:scannet_realworld}.

ScanNet presents a substantially harder upsampling setting than the synthetic benchmarks because the sparse observations are affected by missing regions, non-uniform density, measurement noise, and complex indoor geometry. Although the CD differences among the three methods are relatively modest, the qualitative results reveal a clearer distinction. PUFM and PUFM++ recover the overall scene geometry but tend to lose or smooth fine local structures in difficult regions. In contrast, MRFM reconstructs more complete and visually detailed geometry, indicating that its patch-adaptive refinement remains beneficial under real-world sparsity and noise. These results suggest that the advantage of MRFM is not limited to synthetic test shapes and can transfer to more challenging real-world point-cloud observations.